\documentclass[conference]{IEEEtran}
\IEEEoverridecommandlockouts
\usepackage[T1]{fontenc}
\usepackage{cite}
\usepackage{amsmath,amssymb,amsfonts}
\usepackage{graphicx}
\graphicspath{{figure/}}
\usepackage{textcomp}
\usepackage[table]{xcolor}
\usepackage{booktabs}
\usepackage{multirow}
\usepackage{url}
\usepackage{fontawesome5}
\usepackage{microtype}
\usepackage[normalem]{ulem}
\usepackage[hidelinks]{hyperref}
\makeatletter
\g@addto@macro\normalsize{%
  \setlength{\abovedisplayskip}{5pt plus 1pt minus 1pt}%
  \setlength{\belowdisplayskip}{5pt plus 1pt minus 1pt}%
  \setlength{\abovedisplayshortskip}{3pt plus 1pt minus 1pt}%
  \setlength{\belowdisplayshortskip}{3pt plus 1pt minus 1pt}%
}
\makeatother
\AtBeginDocument{\normalsize\setlength{\jot}{2pt}}
\begin{document}
\title{LEGOSpec: A Composable and Efficient Framework for Speculative Draft Design and Training}
\author{
\IEEEauthorblockN{
Hanzhen Wang$^{1,\dagger}$,
Jiaming Xu$^{1,3,\dagger}$,
Guohao Dai$^{1,2,3,*}$
}
\IEEEauthorblockA{
$^{1}$Shanghai Jiao Tong University \quad
$^{2}$Infinigence AI \quad
$^{3}$SII \\
$^{\dagger}$Equal contribution. \quad
$^{*}$Corresponding author: daiguohao@sjtu.edu.cn
}
}
\maketitle
\begin{abstract}
Speculative decoding accelerates large language model (LLM) generation using a lightweight draft. Existing methods evaluate complete recipes under different training configurations, leaving component effects, interactions, and composition benefits unclear. Therefore, we present \textbf{LEGOSpec}, a composable framework for speculative draft design and training. LEGOSpec decomposes draft models into Context, Backbone, Proposal, and Objective components that share a unified training infrastructure. Using LEGOSpec, we systematically vary components and evaluate selected combinations under matched training conditions to isolate their effects and interactions. These experiments reveal three findings: (1) Different workloads favor different backbones, depending on context length and memory budget, with within-block dependency modeling particularly important for long contexts. (2) Proposal design must balance acceptance and latency. (3) The training objective should match token distributions locally and optimize acceptance length across the block. These findings guide workload-aware configurations and \textit{DFlux}, a balanced general-purpose draft combining a convolution-equipped SWA--GQA hybrid backbone, Residual Corrector, and hierarchical supervision. Across Qwen and Gemma models, \textit{DFlux} improves acceptance length over the strongest baseline by up to 19.7\% on the regular benchmark and 34.7\% on long-context workload. To support efficient training, LEGOSpec integrates metadata-aware scheduling and asynchronous execution, increasing training throughput by up to 79\% and accelerating offline supervision preparation by up to $3.11\times$ over SpecForge.
\end{abstract}
\par
{\normalfont\ttfamily\footnotesize
\noindent\begin{tabular}{@{}ll@{}}
\href{https://github.com/SJTU-DAI-Sys/LEGOSpec}{%
    \raisebox{-0.15em}{\scalebox{1.5}{\faGithub}}\hspace{0.1em}
    \texttt{SJTU-DAI-Sys/LEGOSpec}%
} &
\href{https://huggingface.co/collections/SJTU-DAI-Sys/legospec}{%
    \raisebox{-0.4em}{\includegraphics[height=1.6em]{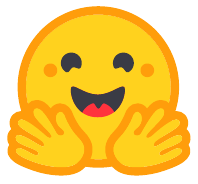}}\hspace{0.1em}
    \texttt{SJTU-DAI-Sys/LEGOSpec}%
} \\
\end{tabular}}
\vspace{2pt}
\vspace{-10pt}
\section{Introduction}

\begin{figure}
    \centering
    \includegraphics[width=01\linewidth]{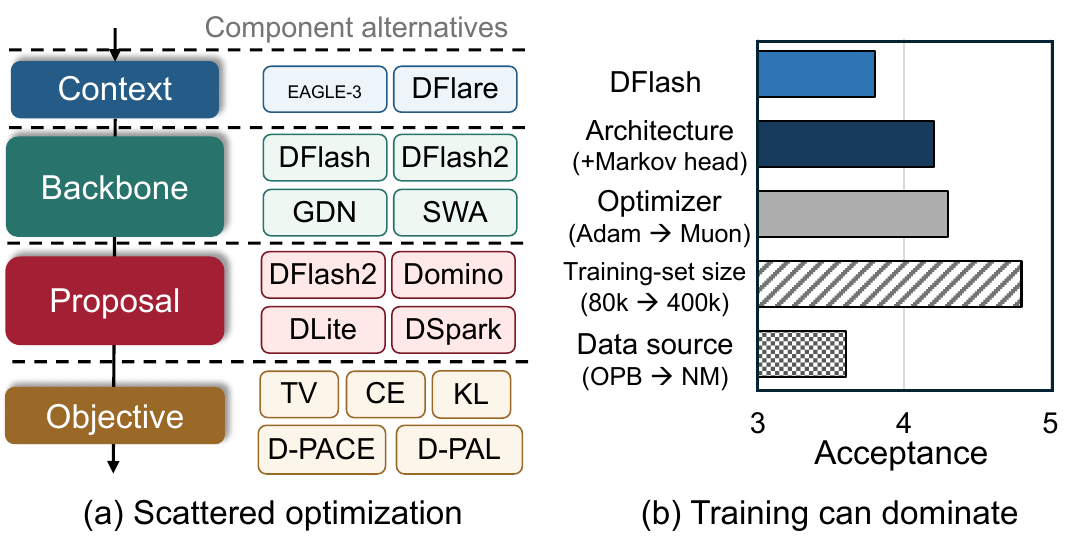}
    \vspace{-15pt}
    \caption{(a) Alternatives in context, backbone, proposal, and objective. (b) In DFlash, changing the optimizer or increasing the training-set size from 80K to 400K can match or exceed the gains from adding a Markov head, while changing the data source can reduce acceptance.}
    \label{fig:draft-motivation}
    \vspace{-10pt}
\end{figure}

Speculative decoding accelerates autoregressive LLMs through parallel target verification of draft tokens~\cite{leviathan2023,chen2023accelerating}. Existing methods improve different draft components~\cite{li2026eagle,wang2025specprune,xu2026specontext,zhang2026dflare,xu2026dlite,samarin2026lk,zhou2024distillspec}, but comparisons of complete recipes under different training configurations obscure component effects, interactions, and composition benefits. Figure~\ref{fig:draft-motivation} shows that training changes can affect acceptance as much as architectural changes. Controlled comparisons are therefore needed to distinguish component effects from differences in training data and optimization.

We present \textbf{LEGOSpec}, a composable framework for speculative draft design and training. LEGOSpec organizes draft recipes along four axes: \textit{Context} selects and fuses target-model features; \textit{Backbone} processes historical context and models dependencies within the draft block; \textit{Proposal} produces candidate distributions, optionally conditioned on preceding tokens; and \textit{Objective} defines the supervision. Using LEGOSpec, we systematically replace individual components around a shared reference and evaluate selected combinations under matched training conditions to isolate component effects and interactions (Section~\ref{sec:component-study}). This shared experimental basis connects the component findings to workload-aware design and to the general-purpose \textit{DFlux} recipe.
The controlled experiments reveal three findings. \textbf{First, different workloads favor different backbones.} Full attention~\cite{vaswani2017attention} favors short-context tasks, whereas sliding-window attention(SWA)~\cite{beltagy2020longformer} and convolution~\cite{dflash2} improve long-context acceptance; compact historical states increase context capacity under memory constraints. \textbf{Second, proposal design should balance acceptance and latency while retaining full-vocabulary outputs. Third, supervision should match token distributions locally and optimize acceptance length across the block.}

These findings guide \textbf{workload-aware draft configurations}: dense full attention achieves the highest \textit{Regular} acceptance, while SWA leads on \textit{Long}. At matched memory budgets, efficient backbones such as SWA and MLA support $5$--$7\times$ the MHA context capacity on Qwen3.5-9B (Section~\ref{sec:workload-design}).

Building on these findings, we also develop \textit{DFlux}, a balanced general-purpose draft combining a convolution-equipped SWA--GQA hybrid backbone, Residual Corrector, and hierarchical supervision. Across Qwen~\cite{yang2025qwen3} and Gemma~\cite{team2026gemma} models, \textit{DFlux} improves acceptance length over the strongest baseline by up to 19.7\% on regular benchmarks and 34.7\% on long-context workloads.

LEGOSpec uses sequence and anchor metadata for lookahead scheduling and cost-aware placement, reducing wasted target computation and draft-worker imbalance. Together with asynchronous execution (Section~\ref{sec:framework}), these optimizations improve training throughput over SpecForge~\cite{li2026specforge} by up to 79\% and accelerate offline supervision preparation by up to $3.11\times$. Our contributions are: (1) A composable framework and controlled study of component effects and interactions under matched conditions. (2) Finding-guided workload-aware configurations and a balanced general-purpose draft. (3) Efficient training infrastructure with metadata-aware scheduling and asynchronous execution.

\section{Background and Related Work}
\subsection{Speculative Decoding Design}
Each round emits an accepted draft prefix and one additional token, preserving the target distribution through rejection sampling~\cite{leviathan2023,chen2023accelerating}. The resulting decoding speedup is approximately
\begin{equation}
\mathrm{Speedup}\approx\frac{\tau\,t_{\mathrm{AR}}}{t_{\mathrm{draft}}+t_{\mathrm{verify}}+t_{\mathrm{other}}},
\end{equation}
where $\tau$ denotes mean acceptance length, including the guaranteed token emitted in each round. Here, $t_{\mathrm{AR}}$ is the mean autoregressive generation time per token, and the denominator sums per-round drafting, verification, and other overheads.

EAGLE-3~\cite{li2026eagle} fuses target features for autoregressive drafting. DFlash~\cite{chen2026dflash} uses block diffusion, while DFlare~\cite{zhang2026dflare} learns layer-specific feature fusion. DSpark~\cite{cheng2026dspark} and DLite~\cite{xu2026dlite} model block dependencies through sequential correction and recurrent heads, respectively. DFlash2~\cite{dflash2} combines candidate selection with block-local convolution.

\subsection{Training Objectives}
DFlash~\cite{chen2026dflash} uses position-decayed CE loss, DART~\cite{liu2026dart} uses KL loss for distillation, and DSpark~\cite{cheng2026dspark} combines token supervision with TV loss for distribution matching. Negative log-acceptance loss (NLA), an LK loss~\cite{samarin2026lk}, directly optimizes token acceptance, while D-PACE~\cite{wu2026d} derives adaptive weights for CE loss from an accepted-length surrogate. D-PAL loss~\cite{wu2026beyond} uses distribution overlap to compute prefix-dependent weights for every position.

\subsection{Draft Model Training}
A draft model uses hidden states obtained from the target model. Token labels or target distributions supervise draft predictions. For block drafting, sampled anchor tokens define prediction blocks~\cite{chen2026dflash}. Online training obtains target features during training, whereas offline training precomputes and caches them. SpecForge~\cite{li2026specforge} decouples target inference from draft training and uses SGLang for feature extraction. This workflow comprises target prefill, supervision delivery, and draft updates.

\section{Controlled Component Study}
\label{sec:component-study}
We vary Context, Backbone, Proposal, and Objective around a shared Qwen3-4B reference trained on 80K samples: DFlash context, five dense GQA layers, RNN proposal, and the Mix loss ($0.1\,\mathrm{CE}+0.9\,\mathrm{TV}$) (Table~\ref{tab:component-overview}). All configurations are evaluated at $T=1$, \textit{Regular} averages the six short-context benchmarks as Section~\ref{sec:experimental-setup}, and \textit{Long} averages six LongBench-v2 domains with 32K--64K-token inputs. Appendix~\ref{app:component-details} gives the objective definitions and training curves.

\subsection{Context}
With 80K training samples, DFlare underperforms DFlash on both \textit{Regular} and \textit{Long}, with a 33.8\% lower acceptance length on \textit{Long} (Table~\ref{tab:context-data-scale}). Since this differs from the publicly reported results~\cite{zhang2026dflare}, we further examine data scaling. With 400K and 1M training samples, DFlare still does not show a consistent advantage over DFlash. We therefore retain DFlash as the default for subsequent comparisons.

DFlare~\cite{zhang2026dflare} supplies each draft layer $\ell$ with a scalar-weighted mixture of target layers:
\begin{equation}
f_t^{(\ell)}=\operatorname{RMSNorm}\!\left(\sum_{j=1}^{M}\operatorname{softmax}(W^{\mathrm{fuse}}_{\ell,:})_j h_t^{(j)}\right).
\end{equation}
where $h_t^{(j)}$ is the hidden state from selected target layer $j$ at position $t$, and $M$ is the number of selected layers. Each scalar fusion weight is shared across channels, so the weighted sum combines corresponding channels across layers without learned cross-channel mixing. DFlash instead projects concatenated target features, mixing feature channels before injection. This distinction offers a possible explanation for the comparison and suggests that layer-specific access alone does not replace channel mixing.

\begin{table}[!t]
\centering
\caption{Context fusion with different training-set sizes.}
\label{tab:context-data-scale}
\footnotesize
\setlength{\tabcolsep}{2pt}
\renewcommand{\arraystretch}{1}
\begin{tabular*}{\columnwidth}{@{\extracolsep{\fill}}lcccccc@{}}
\toprule
Context & \multicolumn{2}{c}{80K} & \multicolumn{2}{c}{400K} & \multicolumn{2}{c}{1M} \\
\cmidrule(lr){2-3}\cmidrule(lr){4-5}\cmidrule(lr){6-7}
 & Regular & Long & Regular & Long & Regular & Long \\
\midrule
DFlash & \textbf{4.472} & \textbf{2.450} & \textbf{4.902} & 2.493 & 5.072 & \textbf{2.512} \\
DFlare & 4.391 & 1.623 & 4.829 & \textbf{2.496} & \textbf{5.082} & 2.428 \\
\bottomrule
\end{tabular*}
\vspace{-10pt}
\end{table}

\subsection{Backbone}
We compare five-layer backbones under comparable parameter budgets (Table~\ref{tab:study-backbone-structure}): multi-head attention (MHA), grouped-query attention (GQA), sliding-window attention (SWA), gated DeltaNet (GDN), and multi-head latent attention (MLA). Historical access and within-block mixing play different roles in draft prediction. Dense MHA and GQA favor \textit{Regular} acceptance, whereas SWA and GDN perform better on \textit{Long} (Table~\ref{tab:component-overview}). Relative to GQA, SWA improves \textit{Long} from 2.450 to 3.023 while reducing \textit{Regular} from 4.472 to 4.422. Adding convolution to GQA improves \textit{Long} by 20\% to 2.950, with only a small change on \textit{Regular}. These differences motivate examining how growing history affects a fixed-size draft block.

\begin{figure}[!t]
\centering
\includegraphics[width=\columnwidth]{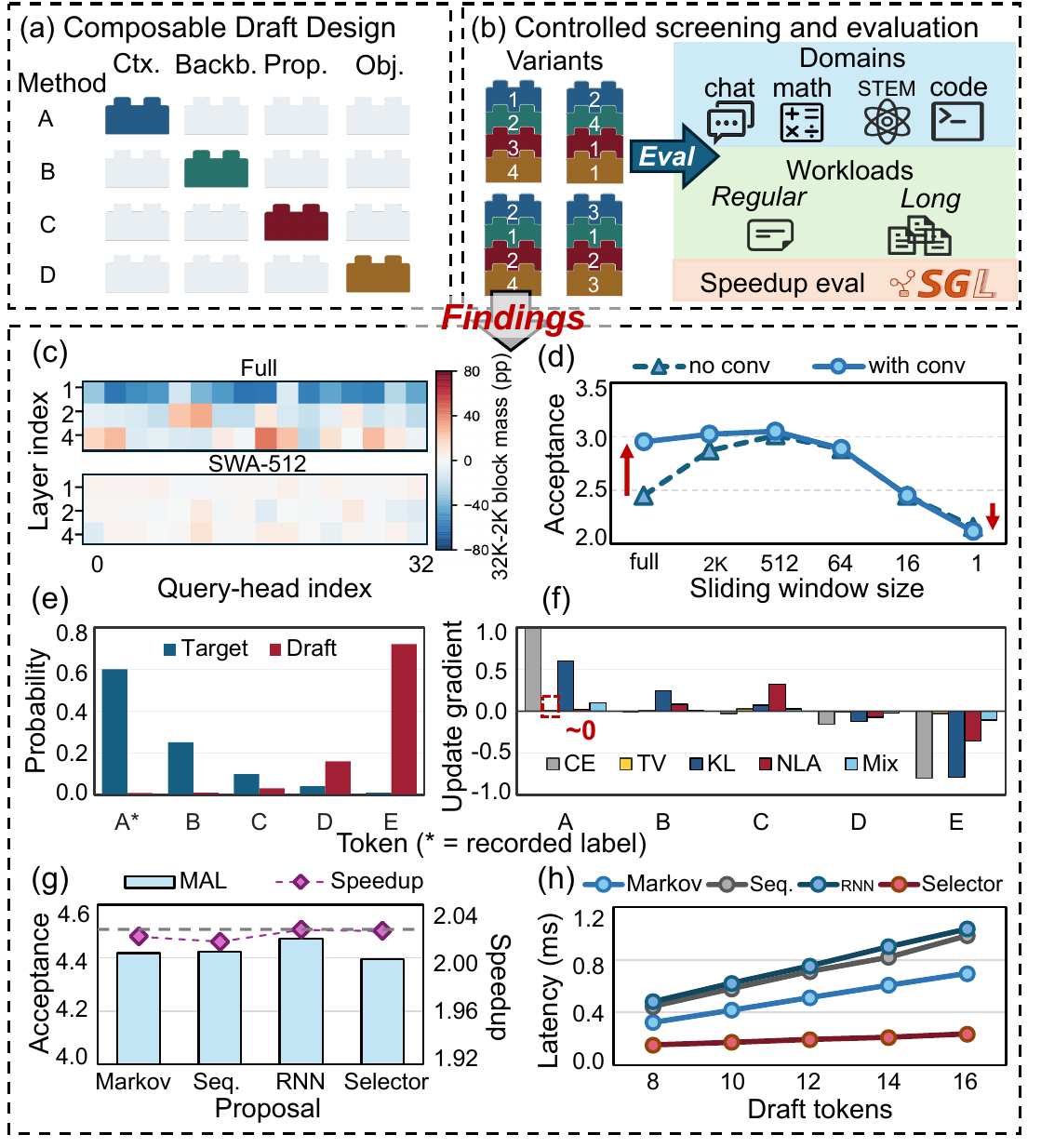}
\vspace{-15pt}
\caption{Controlled component study. (a)--(b) Draft components and matched evaluation. (c) Within-block attention mass from 2K to 32K context. (d) Convolution effects across attention windows. (e)--(f) Token probabilities and negative logit gradients. (g) Proposal acceptance and speedup. (h) Proposal latency versus draft length.}
\label{fig:tech1}
\vspace{-10pt}
\end{figure}

Following DFlash2's attention analysis~\cite{dflash2}, we examine how context growth affects attention within the draft block. Under full attention, the historical key set grows with context length while the draft block remains fixed, increasing the number of historical positions competing for attention mass. This can shift aggregate attention toward history and weaken within-block information exchange. Figure~\ref{fig:tech1}(c) supports this interpretation: from 2K to 32K context, most full-attention query heads allocate less weight to keys within the draft block, whereas SWA-512 remains nearly unchanged. SWA limits this competition by bounding visible history. GDN offers a different mechanism, compressing history into a fixed-size recurrent state rather than extending attention over a growing history.

Convolution directly mixes neighboring block tokens at each layer, providing within-block information exchange independently of attention allocation. If this path compensates for weakened block dependencies, its additional benefit should diminish when SWA already preserves them. The window sweep follows this prediction: convolution's gain decreases as the SWA window narrows and becomes negative at a window of one (Fig.~\ref{fig:tech1}(d)). This interaction supports a shared explanation for their long-context benefits.

Backbones also differ in historical-state storage (Table~\ref{tab:component-overview}). Full-attention caches grow with context length; MLA reduces storage per token, whereas SWA and GDN retain bounded historical states, reducing cache size
by 500$\times$. A smaller draft state leaves more memory available for context under a fixed system memory budget, motivating the maximum-context analysis in Section~\ref{sec:memory-workload}.

\noindent\textbf{Takeaway 1: different workloads favor different backbones.}
Full attention favors short contexts; SWA preserves within-block attention, and convolution strengthens local dependencies. Compact SWA, GDN, and MLA states also enable longer contexts under memory constraints.

\subsection{Proposal}
Dependency correction improves acceptance but also introduces work on the drafting critical path. RNN sequentially updates a hidden state to summarize the sampled token prefix, whereas Markov conditions its correction only on the immediately preceding token. Sequential recurrently predicts the next distribution. RNN achieves the highest acceptance (Table~\ref{tab:component-overview}). However, recurrent prefix processing requires additional computation after each sampling step.

Selector instead computes adjacent candidate transition scores for all positions in parallel, leaving traversal and sampling sequential. The latency and end-to-end measurements expose the tradeoff (Fig.~\ref{fig:tech1}(g)--(h)): Selector has lower proposal latency and reaches speedup close to RNN despite lower acceptance. Thus, a proposal's contribution depends on both how many tokens survive verification and how much time it adds before verification can begin.

These results motivate improving prediction quality while retaining parallel transition scoring. Residual Corrector builds on this structure to support full-vocabulary distribution correction and supervision. Combined with hierarchical supervision, it surpasses RNN in acceptance in our component ablation (Section~\ref{sec:component-ablation}).

\noindent\textbf{Takeaway 2: balance acceptance and latency in dependency correction.}
Proposal design should optimize acceptance and execution latency jointly. Parallel transition scoring retains predecessor conditioning while reducing serial proposal work.

\begin{table}[t]
\centering
\caption{Component-wise acceptance at $T=1$. Cache memory is computed at a context length of 64K tokens.}
\label{tab:component-overview}
\footnotesize
\definecolor{componentBlue}{RGB}{20,65,120}
\setlength{\tabcolsep}{1pt}
\renewcommand{\arraystretch}{1}
\begin{tabular*}{\columnwidth}{@{\extracolsep{\fill}}lllllrrr@{}}
\toprule
Group & Context & Backbone & Proposal & Objective & Regular & Long & \shortstack{Cache\\(MiB)} \\
\midrule
\textbf{Ref.} & \textbf{DFlash} & $\boldsymbol{5\times}$\textbf{GQA} & \textbf{RNN} & \textbf{Mix} & \textbf{4.472} & \textbf{2.450} & \textbf{1280} \\
\midrule
A & DFlare & --- & --- & --- & 4.391 & 1.623 & --- \\
\midrule
B & --- & $5\times$GQA\,{\scriptsize +Conv} & --- & --- & \textit{\textbf{4.489}} & \textit{\textbf{2.950}} & 1280 \\
 & --- & $5\times$MHA & --- & --- & 4.459 & 2.391 & 5120 \\
 & --- & $5\times$SWA & --- & --- & 4.422 & {\textbf{3.023}} & 10.00 \\
 & --- & $5\times$GDN & --- & --- & 4.226 & 2.589 & 10.18 \\
 & --- & $5\times$MLA & --- & --- & 4.436 & 2.384 & 360 \\
\midrule
C & --- & --- & Parallel & --- & 4.100 & 2.337 & --- \\
 & --- & --- & Markov & --- & 4.416 & 2.388 & --- \\
 & --- & --- & Sequential & --- & 4.408 & 2.389 & --- \\
 & --- & --- & Selector* & CE & \textit{\textbf{4.294}} & 2.180 & --- \\
\midrule
D & --- & --- & --- & CE & 4.193 & 2.290 & --- \\
 & --- & --- & --- & TV & 4.131 & 2.320 & --- \\
 & --- & --- & --- & KL & 4.422 & 2.436 & --- \\
 & --- & --- & --- & NLA & 4.421 & 2.439 & --- \\
 & --- & --- & --- & D-PACE & 4.242 & 2.074 & --- \\
 & --- & --- & --- & D-PAL & {\textbf{4.524}} & 2.103 & --- \\
\bottomrule
\end{tabular*}\par
\vspace*{2pt} 
{\scriptsize *The selector retains only the top-$k$ logits; we use CE loss following DFlash2.\par}
\vspace{-10pt}
\end{table}

\subsection{Objective}
\textbf{Token-level supervision.} Distribution matching supervises the draft with target probabilities over the full vocabulary rather than a single token label. Distribution loss NLA, KL and Mix losses improve \textit{Regular} acceptance with label-only CE loss(Table~\ref{tab:component-overview}). Yet standalone TV loss falls behind of CE, though it optimizes the acceptance rate directly.

Figure~\ref{fig:tech1}(e)--(f) illustrates the optimization difficulty in a low-overlap case. For tokens with high target probability but very low draft probability, TV loss produces near-zero gradients with respect to the draft logits, slowing convergence. In Mix loss ($0.1\,\mathrm{CE}+0.9\,\mathrm{TV}$), the CE term strengthens corrective gradients in this regime, while the TV term encourages distribution alignment. The illustrated case shows that Mix loss combines stronger gradients at low overlap with close alignment at convergence, explaining its advantage over standalone TV loss. Objective definitions are in Appendix~\ref{app:objective-definitions}.

\textbf{Block-level supervision.} A later token contributes to acceptance length only if the preceding prefix is accepted. D-PACE and D-PAL losses account for this dependency by adaptively weighting local CE and KL losses. On \textit{Regular}, both outperform their corresponding local losses, supporting acceptance-aware aggregation. Although D-PAL loss underperforms on \textit{Long}, adding convolution, which benefits long-context drafting (Section~\ref{sec:component-ablation}), allows it to outperform Mix loss on both \textit{Regular} and \textit{Long}. This suggests that complementary components can work together to improve performance across workloads.

\noindent\textbf{Takeaway 3: match distributions locally and optimize acceptance length across the block.}
Proposals should retain full-vocabulary outputs for distributional supervision, while the final objective should account for prefix acceptance when aggregating local losses. These two levels motivate joint proposal--objective design and hierarchical supervision.

\section{Finding-Guided Workload-Aware Design}
\label{sec:workload-design}
\subsection{Context-Length-Dependent Backbone Selection}
\begin{figure}[t]
    \centering
    \includegraphics[width=\columnwidth]{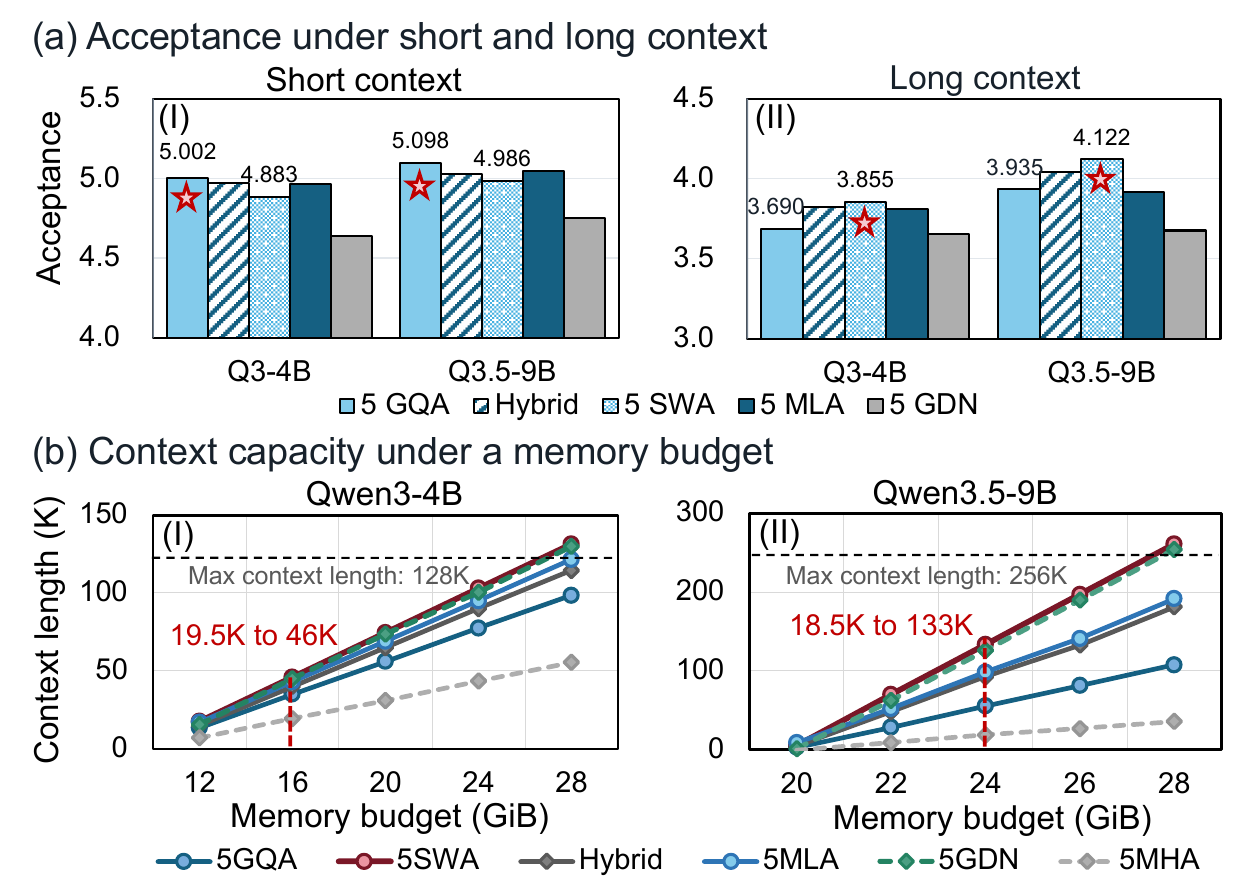}
    \vspace{-15pt}
    \caption{(a) Acceptance of different backbones with matched convolution, RNN proposal, and Mix loss. (b) Maximum context length under each memory budget, including weights and runtime overhead.}
    \vspace{-10pt}
    \label{fig:workload-acceptance}
\end{figure}

With 400K training samples, dense GQA leads on \textit{Regular} and SWA leads on \textit{Long} for both target models (Fig.~\ref{fig:workload-acceptance}(a)). On Qwen3-4B, their respective advantages are 2.44\% and 4.47\%. These results favor full historical access for short contexts and bounded historical KV access for long-context drafting; the SWA--GQA hybrid balances these preferences.

\subsection{Maximum Context Length under Memory Constraints}
\label{sec:memory-workload}
Compact historical states increase context capacity under the same total runtime memory budget (Fig.~\ref{fig:workload-acceptance}(b)). Replacing five dense MHA layers with SWA extends Qwen3-4B from 19.5K to 46K tokens at 16~GiB and Qwen3.5-9B from 18.5K to 133K at 24~GiB (\(7.19\times\)). MLA and the hybrid provide intermediate capacity, while GDN also supports substantially longer contexts. SWA and MLA maintain acceptance length comparable to full attention, whereas GDN incurs a larger reduction. These results show that compact backbones can enable longer-context workloads under memory constraints while retaining competitive acceptance length.

\begin{figure}
    \centering
    \includegraphics[width=0.98\linewidth]{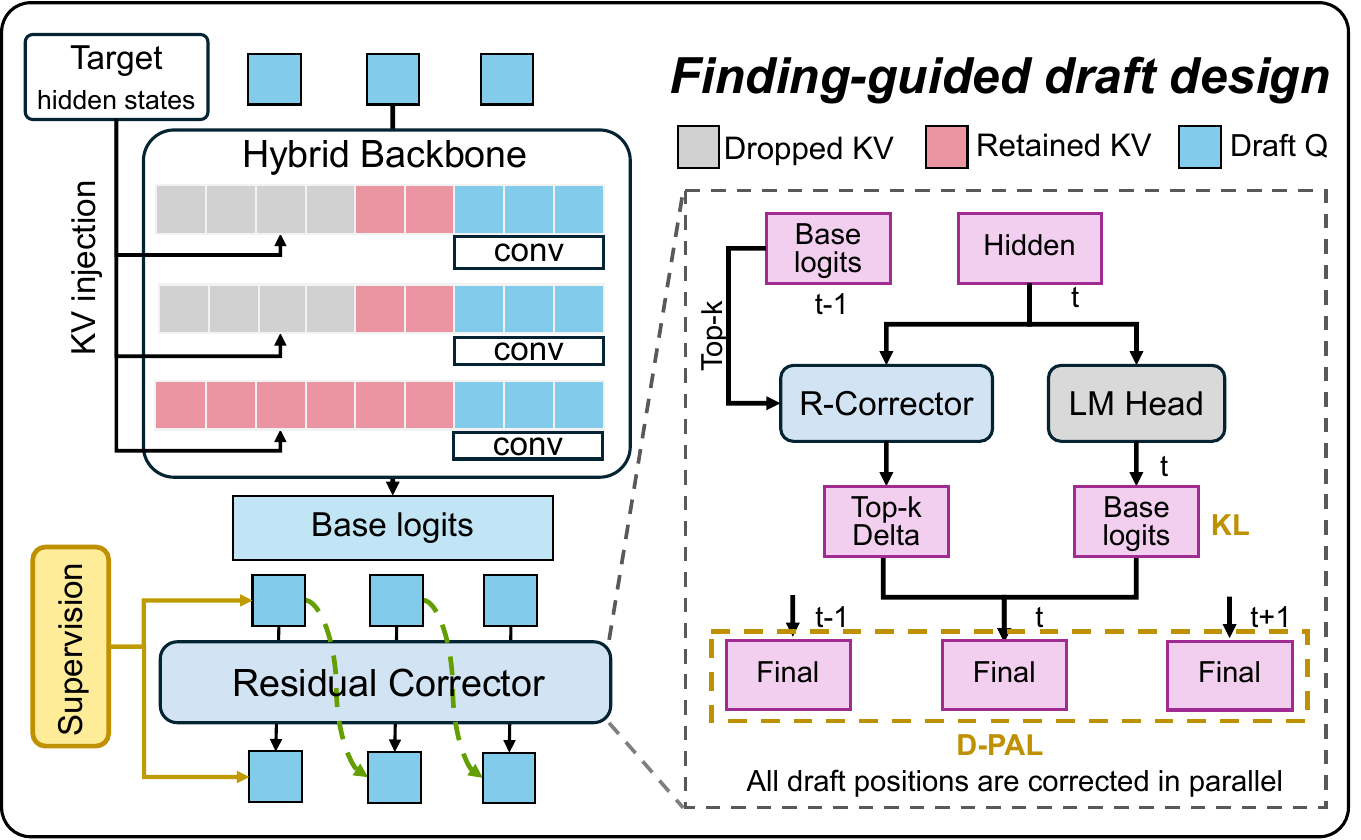}
    \caption{Our finding-guided general-purpose draft combines a hybrid backbone with Residual Corrector and Hierarchical supervision.}
    \vspace{-10pt}
\end{figure}

\subsection{A Balanced General-Purpose Draft}
\label{sec:dflux}
We combine three refinements into a general-purpose draft \textit{DFlux} that balances short- and long-context acceptance and historical-state storage.

\textbf{SWA--GQA hybrid backbone.} Deeper full-attention layers are less sensitive to context length. We use four SWA layers followed by one dense GQA layer, with block-local convolution. SWA preserves within-block attention, convolution mixes neighboring draft positions, and the final GQA layer incorporates full history. Only that final layer retains a context-growing cache. This hybrid improves \textit{Long} acceptance while maintaining \textit{Regular} performance (Fig.~\ref{fig:workload-acceptance}).


\textbf{Residual Corrector (R-Corrector).} To preserve Selector's parallelism while supporting full-vocabulary supervision, we propose Residual Corrector. Our measurements show that the top-16 candidates account for 99\% of the target--draft distribution overlap. We therefore restrict correction to these few candidates and retain the backbone's base logits for the rest of the vocabulary, reducing correction work to a small candidate space.

Let $h_i$ and $z_i^{\mathrm{base}}$ denote the backbone hidden state and logits, and let $C_i$ contain the top-$k$ candidates. For predecessor token $a$ and candidate $v$, the correction and final logits are
\begin{equation}
\begin{aligned}
\Delta_i(a,v)&=\big(E^{-}_{a}\odot Wh_i\big)^\top E^{+}_{v},\\
z_i^{\mathrm{final}}(v\mid a)&=z_i^{\mathrm{base}}(v)
+\mathbf{1}_{v\in C_i}\,\Delta_i(a,v),
\end{aligned}
\end{equation}
where $W$ projects the hidden state to $r$ dimensions and [$E^{-},E^{+}$] are predecessor and successor codebooks. For each position, scoring all pairs in $C_{i-1}\times C_i$ produces a $k\times k$ correction table. These tables are computed for all positions in parallel. If the predecessor token $a$ is outside the candidate set $a\notin C_{i-1}$, we set the correction $\Delta_i(a,v)=0$ to zero.  With rejection sampling, the final output theoretically preserves the target distribution (Appendix~\ref{app:sampling-correctness}).

\textbf{Hierarchical supervision.} The backbone output aims to include target tokens in the top-$k$ candidate set at each position, while the final output aims to maximize acceptance length. We therefore supervise them separately: auxiliary forward KL loss encourages the backbone distribution to cover plausible target tokens, while final-output D-PAL loss uses prefix-dependent aggregation to optimize acceptance length across the draft block. Both branches use the full vocabulary. The total loss is
\begin{equation}
\mathcal{L}_{\mathrm{train}}
= \lambda_{\mathrm{aux}}\,\mathcal{L}_{\mathrm{KL}}(p,q^{\mathrm{base}}) + \mathcal{L}_{\mathrm{D\text{-}PAL}}(p,q^{\mathrm{final}}),
\end{equation}
Here, $p$ is the target distribution, and the auxiliary KL is averaged over valid draft positions. We set $\lambda_{\mathrm{aux}}=0.2$ to balance loss scales.

\section{LEGOSpec Framework Design}
\label{sec:framework}
\subsection{Model Assembly and Unified Training}
LEGOSpec connects Context, Backbone, Proposal, and Objective to a shared trainer through a recipe adapter. The adapter supplies legal anchor positions and supervision layouts, allowing component replacements without changing the training loop. Target workers provide the required hidden states directly for online training or store them for offline training. For the same target model and tokenized training data, recipes whose required target layers are cached can reuse the supervision. This separates draft design from target feature extraction and supports controlled comparisons across component combinations.

\begin{figure}[t]
    \centering
    \includegraphics[width=\columnwidth]{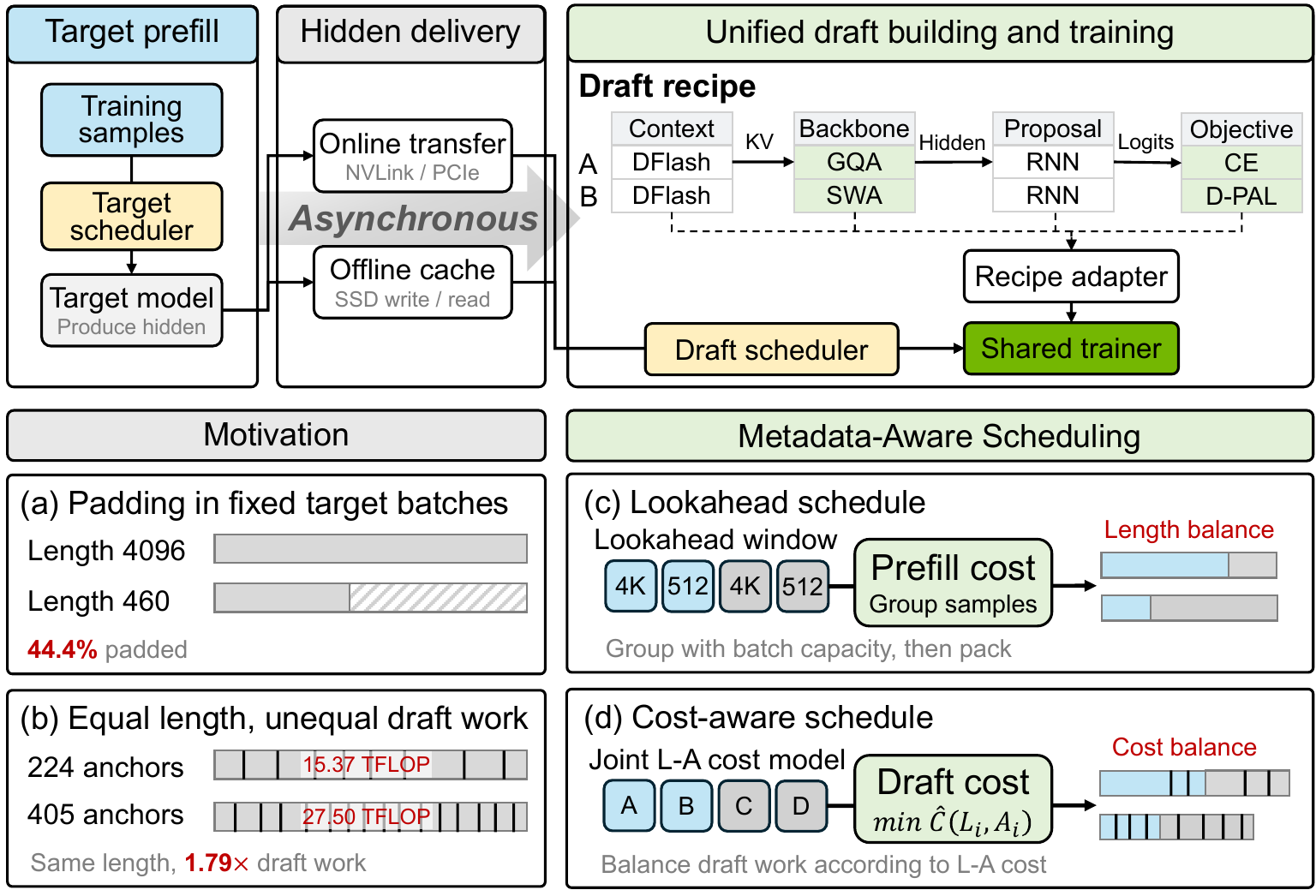}
    \vspace{-15pt}
    \caption{LEGOSpec pipeline (top) and metadata-aware scheduling (bottom).}
    \vspace{-5pt}
    \label{fig:legospec-framework}
\end{figure}

\subsection{Metadata-Aware Scheduling}
\textbf{Motivation.} Existing frameworks reuse serving engines such as SGLang~\cite{zheng2024sglang} and vLLM~\cite{kwon2023efficient} for target prefill. Scheduling from the current request queue limits cross-batch planning when training submits one batch at a time. Training sequences, however, have known lengths and supervision masks, enabling lookahead grouping and packed execution. Draft scheduling faces a different imbalance: equal-length samples can have different anchor counts and therefore different computation costs. Figure~\ref{fig:legospec-framework} shows a $1.79\times$ draft cost difference, which length-only scheduling cannot capture.

\begin{table*}[!t]
\centering
\caption{Mean acceptance length $\tau$ at $T=1$ on short- and long-context workloads, averaged over five random seeds.}
\label{tab:baseline-comparison}
\setlength{\tabcolsep}{1.5pt}
\newcommand{\datasetheading}[1]{\makebox[0.075\textwidth][c]{#1}}
\renewcommand{\arraystretch}{1.0}
\begin{tabular*}{0.99\textwidth}{@{\extracolsep{\fill}}cccccccccccc@{}}
\toprule
 & & \multicolumn{6}{c}{Short-context workloads} & \multicolumn{3}{c}{LongBench-v2 domains} & Long coding \\
\cmidrule(lr){3-8}\cmidrule(lr){9-11}\cmidrule(lr){12-12}
Model & Method & \datasetheading{Alpaca} & \datasetheading{MT-Bench}
 & \datasetheading{GSM8K} & \datasetheading{MATH-500}
 & \datasetheading{MBPP} & \datasetheading{LCB}
 & \datasetheading{MD-QA} & \datasetheading{L-ICL}
 & \datasetheading{SD-QA} & \datasetheading{SWE} \\
\midrule
\multirow{5}{*}{Qwen3-4B} & DFlash & 3.23 & 3.75 & 5.63 & 5.32 & 4.77 & 3.97 & 2.14 & 2.52 & 2.50 & 1.58 \\
 & DFlash2 & 3.25 & 3.88 & 5.87 & 5.64 & 4.83 & 3.89 & 3.12 & 3.45 & 3.17 & 2.96 \\
 & DSpark & 3.57 & 3.89 & 6.04 & 5.61 & 5.12 & 4.30 & 2.42 & 2.75 & 2.67 & 2.24 \\
& DLite & 3.56 & 3.88 & 6.12 & 5.98 & 5.14 & 4.04 & 2.74 & 3.11 & 2.98 & 3.03 \\
 & \textbf{DFlux} & \textbf{4.03} & \textbf{4.39} & \textbf{6.47} & \textbf{6.35} & \textbf{5.74} & \textbf{4.95} & \textbf{3.60} & \textbf{3.81} & \textbf{3.63} & \textbf{4.08} \\
\midrule
\multirow{4}{*}{\shortstack[c]{Qwen3.6-\\35B-A3B}} & DFlash & 3.26 & 3.82 & 5.77 & 5.82 & 5.66 & 4.55 & 3.25 & 3.65 & 3.09 & 3.15 \\
 & DFlash2 & 3.45 & 4.00 & 6.06 & 5.99 & 5.88 & 4.66 & 3.48 & 4.05 & 3.34 & 3.62 \\
 & DSpark & 3.39 & 3.75 & 5.66 & 5.85 & 4.97 & 4.11 & 3.14 & 3.76 & 3.17 & 2.84 \\
 & \textbf{DFlux} & \textbf{4.13} & \textbf{4.59} & \textbf{6.89} & \textbf{6.54} & \textbf{6.24} & \textbf{5.09} & \textbf{3.60} & \textbf{4.16} & \textbf{3.58} & \textbf{3.98} \\
\midrule
\multirow{4}{*}{\shortstack[c]{Gemma4\\-12B-it}} & DFlash & 3.36 & 3.69 & 5.94 & 5.54 & 4.80 & 4.14 & 2.72 & 2.78 & 2.65 & 3.39 \\
 & DFlash2 & 3.72 & 4.05 & 6.33 & 6.01 & 5.28 & 4.65 & 2.60 & 2.81 & 2.58 & 3.44 \\
 & DSpark & 3.60 & 3.97 & 6.21 & 5.92 & 5.19 & 4.50 & 3.00 & 3.05 & 3.01 & 4.00 \\
 & \textbf{DFlux} & \textbf{4.12} & \textbf{4.43} & \textbf{6.65} & \textbf{6.42} & \textbf{5.77} & \textbf{5.14} & \textbf{4.02} & \textbf{4.09} & \textbf{3.85} & \textbf{4.52} \\
\bottomrule
\end{tabular*}
\vspace{-10pt}
\end{table*}

\textbf{Target-side lookahead scheduling.} LEGOSpec groups upcoming requests by sequence length within a lookahead window, reducing padding and balancing target-worker prefill work. Packed execution processes valid tokens while retaining sequence boundaries and positions. The resulting hidden states are transferred to draft workers for online training or cached for offline training.

\textbf{Draft-side cost-aware scheduling.} LEGOSpec combines a draft cost model with draft lookahead to group samples and assign work across draft workers.

\textit{Draft cost model.} The planner computes work counts from sequence lengths, anchor layouts, and supervision masks. With sequence length $L$, $A$ anchors, and block size $B$, the query count is $Q=AB$, and full attention has $P=B\sum_{j=1}^{A}t_j+AB^2$ visible Q--K pairs, where $t_j$ is anchor $j$'s history length. The main operation-count proxies are
\begin{equation}
\begin{aligned}
C_{\mathrm{attn}}&\approx(L+Q)H^2+PH,\\
C_{\mathrm{FFN}}&\approx3QHF,\qquad
C_{\mathrm{vocab}}\approx R_{\mathrm{loss}}HV,
\end{aligned}
\end{equation}
where $H$, $F$, and $V$ are hidden, FFN, and vocabulary dimensions, and $R_{\mathrm{loss}}$ counts active loss rows. For SWA, Q--K pairs are counted over window-limited history. For GDN, recurrent-state updates replace the Q–K interaction term. MLA uses the latent dimension to estimate projection costs. Summing per-layer costs with context, proposal, and applicable teacher-projection costs gives sample weight $\widehat C_i$.

For a target batch $\mathcal B$ and $R$ draft ranks, sample assignment minimizes the maximum estimated rank cost:
\begin{equation}
    \begin{aligned}
    \min_{\{\mathcal S_r\}_{r=1}^{R}}\quad &
    \max_{1\le r\le R}
    \sum_{i\in\mathcal S_r}\widehat C_i \\
    \text{s.t.}\quad & \biguplus_{r=1}^{R}\mathcal S_r=\mathcal B.
    \end{aligned}
\end{equation}
Here, $\mathcal S_r$ contains samples assigned to rank $r$. LEGOSpec uses a deterministic greedy algorithm: samples are sorted by cost, grouped into synchronized rounds, and assigned to the least-loaded rank with a free batch slot.

\textit{Draft lookahead.} Applying the draft cost model across upcoming updates provides more choices for balanced grouping and placement. LEGOSpec retains the window's sample occurrences and planned anchors, but can regroup samples across updates. We use a 16-update lookahead window in the framework efficiency experiment.

\subsection{Asynchronous Supervision Pipeline}
LEGOSpec overlaps CPU batch preparation and supervision I/O with GPU computation. Background workers construct anchor layouts, estimate costs, and assemble upcoming batches while draft workers train on the current batch.

Target workers prefill the next batch while completed hidden states are transferred to draft workers for online training or written to storage for offline supervision preparation. During offline training, draft workers prefetch cached hidden states and prepare the next batch while GPU computation proceeds. Bounded queues and reusable buffers limit outstanding transfers and writes. This asynchronous supervision pipeline hides preparation, transfer, and storage latency from training execution.

\section{Evaluation}

\subsection{Experimental Setup}\label{sec:experimental-setup}
\textbf{Training setup and baselines.} We train \textit{DFlux} on 1.4M samples: 900K sampled from Open-PerfectBlend~\cite{xu2024perfect}, 300K sampled from Nemotron-Post-Training-Dataset-v2~\cite{NemotronPostTrainingDatasetV2}, and 200K sampled from WildChat~\cite{zhao2024wildchat}. We retain only the questions and generate responses at $T=1$ to construct the training data. We train for 6 epochs on 16 NVIDIA H800 GPUs with a global batch size of 128. The learning rates for Adam~\cite{kingma2017adammethodstochasticoptimization} and Muon~\cite{liu2025muon} are $6\times10^{-4}$ and $0.01$. We use baseline checkpoints released with the original papers or provided by deepseek-ai and RedHatAI (Appendix~\ref{sec:checkpoint-sources}). We train DFlash2 for Gemma4-12B-it ourselves since no public checkpoint was available. 

\textbf{Evaluation protocol.} We use block size 8 for all models except the Qwen3-4B DFlash2 checkpoint, which retains its native block size of 9. Acceptance evaluations use $T=1$, a maximum generation length of 1,024 tokens, and rejection sampling. We repeat each acceptance benchmark five times with different random seeds and report the average acceptance length to reduce sampling variability.

\textbf{Benchmarks.} Short-context workloads include Chat: Alpaca~\cite{alpaca} and MT-Bench~\cite{zheng2023judging}; Math: GSM8K~\cite{cobbe2021training} and MATH-500~\cite{math500}; Code: MBPP~\cite{austin2021program} and LiveCodeBench v6 (LCB)~\cite{jain2025livecodebench}. Long-context workloads comprise the single-document QA (SD-QA), multi-document QA (MD-QA), and long in-context learning (L-ICL) domains of LongBench-v2~\cite{bai2025longbench}, together with SWE-bench~\cite{jimenez2024swe} prompts based on real repository-level software issues. We select inputs between 32K and 64K tokens from both long-context benchmarks.

\subsection{Comparison with Existing Methods}
We compare \textit{DFlux} with DFlash~\cite{chen2026dflash}, DFlash2~\cite{dflash2}, DSpark~\cite{cheng2026dspark}, and DLite~\cite{xu2026dlite} on Qwen3-4B, Qwen3.6-35B-A3B, and Gemma4-12B-it. \textit{DFlux} achieves the highest acceptance length across all evaluated workloads and target models (Table~\ref{tab:baseline-comparison}), improving over the strongest baseline by up to 34.7\% on SWE-bench and 19.7\% on Alpaca. These gains across chat, math, code, and long-context workloads demonstrate the effectiveness and generalizability of our component refinements.

\subsection{Comparison under Matched Training Conditions}
\begin{figure}[!ht]
    \centering
    \vspace{-5pt}
    \includegraphics[width=1\linewidth]{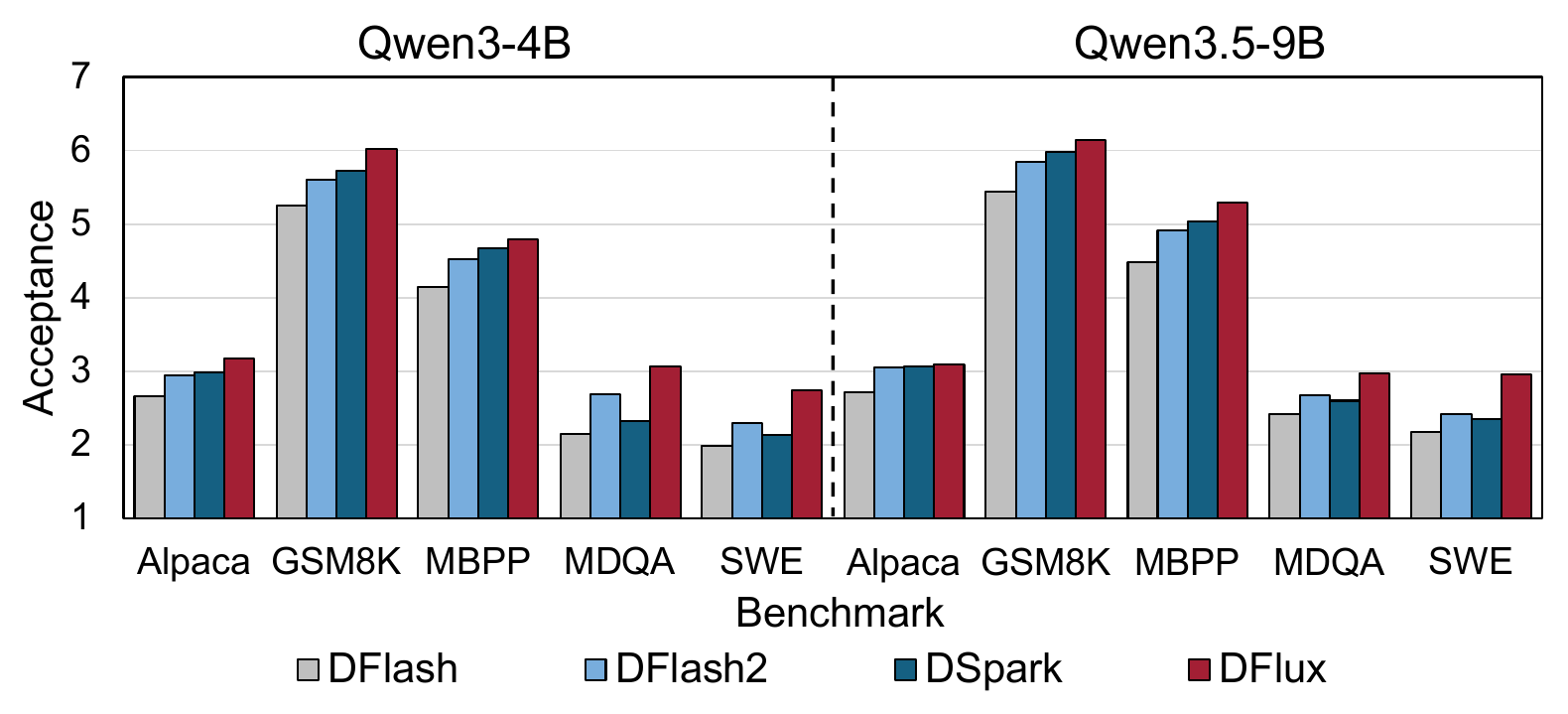}
    \vspace{-20pt}
    \caption{Mean acceptance length $\tau$ at $T=1$ under identical training data and shared training hyperparameters on Qwen3-4B and Qwen3.5-9B.}
    \vspace{-10pt}
    \label{fig:samedata}
\end{figure}

To isolate the benefit of draft design, we train DFlash, DFlash2, DSpark, and \textit{DFlux} on the same 80K Open-PerfectBlend samples with shared training hyperparameters. \textit{DFlux} achieves the highest acceptance on every benchmark for both target models (Fig.~\ref{fig:samedata}), showing that its gains persist under matched training conditions. Detailed acceptance lengths for individual datasets are reported in Table~\ref{tab:matched-training-earlier} in Appendix~\ref{app:matched-training}.

\subsection{Component Ablation}
\label{sec:component-ablation}
Table~\ref{tab:component-combinations} evaluates the backbone, proposal, and supervision refinements on Qwen3-4B and Qwen3.5-9B.

\begin{table}[t]
\centering
\caption{Component ablation with DFlash context.}
\label{tab:component-combinations}
\footnotesize
\setlength{\tabcolsep}{2pt}
\renewcommand{\arraystretch}{1.06}
\begin{tabular*}{\columnwidth}{@{\extracolsep{\fill}}llllrr@{}}
\toprule
Model & Backbone & Proposal & Objective & Regular & Long \\
\midrule
\multirow{6}{*}{Q3-4B} & \uline{GQA} & RNN & Mix & 4.472 & 2.450 \\
 & \uline{Hybrid} & \uline{RNN} & Mix & 4.480 & 3.051 \\
 & Hybrid & \uline{Selector} & CE & 4.392 & 2.927 \\
 & Hybrid & \uline{R-Corrector} & \uline{Mix} & 4.535 & 3.045 \\
 & Hybrid & R-Corrector & \uline{D-PAL} & 4.599 & 3.109 \\
 & Hybrid & R-Corrector & \uline{Hierarchical}  & \textbf{4.618} & \textbf{3.123} \\
\midrule
\multirow{6}{*}{Q3.5-9B} & \uline{GQA} & RNN & Mix & 4.548 & 2.649 \\
 & \uline{Hybrid} & \uline{RNN} & Mix & 4.560 & 3.036 \\
 & Hybrid & \uline{Selector} & CE & 4.422 & 2.875 \\
 & Hybrid & \uline{R-Corrector} & \uline{Mix} & 4.578 & 3.032 \\
 & Hybrid & R-Corrector & \uline{D-PAL} & 4.666 & 3.110 \\
 & Hybrid & R-Corrector & \uline{Hierarchical}  & \textbf{4.687} & \textbf{3.121} \\
\bottomrule
\end{tabular*}
\vspace{-5pt}
\end{table}

\textbf{Backbone.} With RNN and Mix loss fixed, Hybrid improves \textit{Long} acceptance over dense GQA by 24.5\% on Qwen3-4B and 14.6\% on Qwen3.5-9B, with slight \textit{Regular} gains. SWA limits historical-KV competition, while block-local convolution strengthens within-block dependencies.

\textbf{Proposal.} With Mix loss, R-Corrector improves \textit{Regular} acceptance over RNN with nearly unchanged \textit{Long} acceptance. R-Corrector with Mix loss also outperforms Selector with CE loss on both workloads, supporting full-vocabulary distributional supervision while retaining parallel transition scoring.

\textbf{Objective.} Hierarchical supervision achieves the highest acceptance on both models and workloads, outperforming Mix and D-PAL losses. Auxiliary KL loss promotes backbone candidate coverage, while final-output D-PAL loss optimizes acceptance length; their combination improves the final draft.

\subsection{High-concurrency Throughput}
\label{sec:sglang-baselines}
We efficiently implement \textit{DFlux} with SGLang and compare throughput at $T=0$ (Table~\ref{tab:sglang-baselines}). Avg.~$\tau$ denotes mean acceptance length. On Qwen3.6-35B-A3B and Gemma4-12B-it, \textit{DFlux} achieves the highest throughput on all three tasks at every batch size, with $1.57$--$5.62\times$ speedups over AR and 4.7--24.2\% higher throughput than the strongest speculative baseline. Higher acceptance lengths reflect improved prediction accuracy, while parallel transition scoring in R-Corrector reduces sequential proposal work. Together, they enable \textit{DFlux} to improve both decoding latency at batch size one and throughput at larger batch sizes.

\begin{table}[t]
\centering
\caption{Throughput and acceptance with SGLang backbend.}
\label{tab:sglang-baselines}
\footnotesize
\setlength{\tabcolsep}{1pt}
\renewcommand{\arraystretch}{1.12}
\newcommand{\tpsvalue}[1]{{\sbox0{#1}\makebox[\wd0][c]{\scalebox{0.93}{#1}}}}
\newcommand{\tpspair}[2]{\tpsvalue{#1#2$\times$}}
\resizebox{\columnwidth}{!}{%
\begin{tabular}{@{}llccccc@{}}
\toprule
\multirow{2}{*}{Task} & \multirow{2}{*}{Method} & \multicolumn{4}{c}{Batch size (TPS / Speedup)} & \multirow{2}{*}{Avg. $\tau$} \\
\cmidrule(lr){3-6}
 & & 1 & 8 & 32 & 64 & \\
\midrule
\multicolumn{7}{l}{\textbf{Qwen3.6-35B-A3B}} \\
\midrule
\multirow{4}{*}{Alpaca} & AR & \tpspair{131}{/1.00} & \tpspair{1033}{/1.00} & \tpspair{3740}{/1.00} & \tpspair{6113}{/1.00} & \tpsvalue{---} \\
 & DFlash2 & \tpspair{298}{/2.27} & \tpspair{2001}{/1.94} & \tpspair{5817}{/1.56} & \tpspair{7820}{/1.28} & \tpsvalue{3.37} \\
 & DSpark & \tpspair{303}{/2.30} & \tpspair{2005}{/1.94} & \tpspair{5137}{/1.37} & \tpspair{7193}{/1.18} & \tpsvalue{3.26} \\
 & \textbf{DFlux} & \tpspair{\textbf{353}}{\textbf{/2.69}} & \tpspair{\textbf{2336}}{\textbf{/2.26}} & \tpspair{\textbf{7125}}{\textbf{/1.91}} & \tpspair{\textbf{9712}}{\textbf{/1.59}} & \tpsvalue{\textbf{3.85}} \\
\midrule
\multirow{4}{*}{GSM8K} & AR & \tpspair{132}{/1.00} & \tpspair{1033}{/1.00} & \tpspair{3740}{/1.00} & \tpspair{5988}{/1.00} & \tpsvalue{---} \\
 & DFlash2 & \tpspair{517}{/3.92} & \tpspair{3497}{/3.38} & \tpspair{10127}{/2.71} & \tpspair{13281}{/2.22} & \tpsvalue{5.96} \\
 & DSpark & \tpspair{504}{/3.82} & \tpspair{3349}{/3.24} & \tpspair{8552}{/2.29} & \tpspair{11611}{/1.94} & \tpsvalue{5.52} \\
 & \textbf{DFlux} & \tpspair{\textbf{602}}{\textbf{/4.57}} & \tpspair{\textbf{3994}}{\textbf{/3.86}} & \tpspair{\textbf{12058}}{\textbf{/3.22}} & \tpspair{\textbf{16013}}{\textbf{/2.67}} & \tpsvalue{\textbf{6.66}} \\
\midrule
\multirow{4}{*}{MBPP} & AR & \tpspair{132}{/1.00} & \tpspair{1033}{/1.00} & \tpspair{3565}{/1.00} & \tpspair{5677}{/1.00} & \tpsvalue{---} \\
 & DFlash2 & \tpspair{500}{/3.80} & \tpspair{3374}{/3.27} & \tpspair{9255}{/2.60} & \tpspair{11938}{/2.10} & \tpsvalue{5.73} \\
 & DSpark & \tpspair{440}{/3.34} & \tpspair{2916}{/2.82} & \tpspair{7118}{/2.00} & \tpspair{9443}{/1.66} & \tpsvalue{4.77} \\
 & \textbf{DFlux} & \tpspair{\textbf{532}}{\textbf{/4.04}} & \tpspair{\textbf{3533}}{\textbf{/3.42}} & \tpspair{\textbf{10136}}{\textbf{/2.84}} & \tpspair{\textbf{13120}}{\textbf{/2.31}} & \tpsvalue{\textbf{5.88}} \\
\midrule
\multicolumn{7}{l}{\textbf{Gemma4-12B-it}} \\
\midrule
\multirow{4}{*}{Alpaca} & AR & \tpspair{47}{/1.00} & \tpspair{381}{/1.00} & \tpspair{1480}{/1.00} & \tpspair{3168}{/1.00} & \tpsvalue{---} \\
 & DFlash2 & \tpspair{109}{/2.32} & \tpspair{943}{/2.48} & \tpspair{3195}{/2.16} & \tpspair{4122}{/1.30} & \tpsvalue{3.02} \\
 & DSpark & \tpspair{123}{/2.61} & \tpspair{1011}{/2.65} & \tpspair{3408}{/2.30} & \tpspair{4393}{/1.39} & \tpsvalue{3.41} \\
 & \textbf{DFlux} & \tpspair{\textbf{137}}{\textbf{/2.91}} & \tpspair{\textbf{1151}}{\textbf{/3.02}} & \tpspair{\textbf{3833}}{\textbf{/2.59}} & \tpspair{\textbf{4963}}{\textbf{/1.57}} & \tpsvalue{\textbf{3.61}} \\
\midrule
\multirow{4}{*}{GSM8K} & AR & \tpspair{47}{/1.00} & \tpspair{382}{/1.00} & \tpspair{1492}{/1.00} & \tpspair{3253}{/1.00} & \tpsvalue{---} \\
 & DFlash2 & \tpspair{212}{/4.50} & \tpspair{1838}{/4.81} & \tpspair{6394}{/4.29} & \tpspair{8399}{/2.58} & \tpsvalue{5.72} \\
 & DSpark & \tpspair{236}{/5.02} & \tpspair{1950}{/5.10} & \tpspair{6741}{/4.52} & \tpspair{8856}{/2.72} & \tpsvalue{6.29} \\
 & \textbf{DFlux} & \tpspair{\textbf{256}}{\textbf{/5.42}} & \tpspair{\textbf{2149}}{\textbf{/5.62}} & \tpspair{\textbf{7362}}{\textbf{/4.93}} & \tpspair{\textbf{9719}}{\textbf{/2.99}} & \tpsvalue{\textbf{6.56}} \\
\midrule
\multirow{4}{*}{MBPP} & AR & \tpspair{47}{/1.00} & \tpspair{378}{/1.00} & \tpspair{1465}{/1.00} & \tpspair{3114}{/1.00} & \tpsvalue{---} \\
 & DFlash2 & \tpspair{160}{/3.43} & \tpspair{1379}{/3.65} & \tpspair{4614}{/3.15} & \tpspair{5876}{/1.89} & \tpsvalue{4.51} \\
 & DSpark & \tpspair{190}{/4.06} & \tpspair{1555}{/4.12} & \tpspair{5159}{/3.52} & \tpspair{6561}{/2.11} & \tpsvalue{5.25} \\
 & \textbf{DFlux} & \tpspair{\textbf{207}}{\textbf{/4.44}} & \tpspair{\textbf{1739}}{\textbf{/4.60}} & \tpspair{\textbf{5689}}{\textbf{/3.88}} & \tpspair{\textbf{7288}}{\textbf{/2.34}} & \tpsvalue{\textbf{5.55}} \\
\bottomrule
\end{tabular}
}
\end{table}

\begin{figure}[t]
    \centering
    \includegraphics[width=\linewidth]{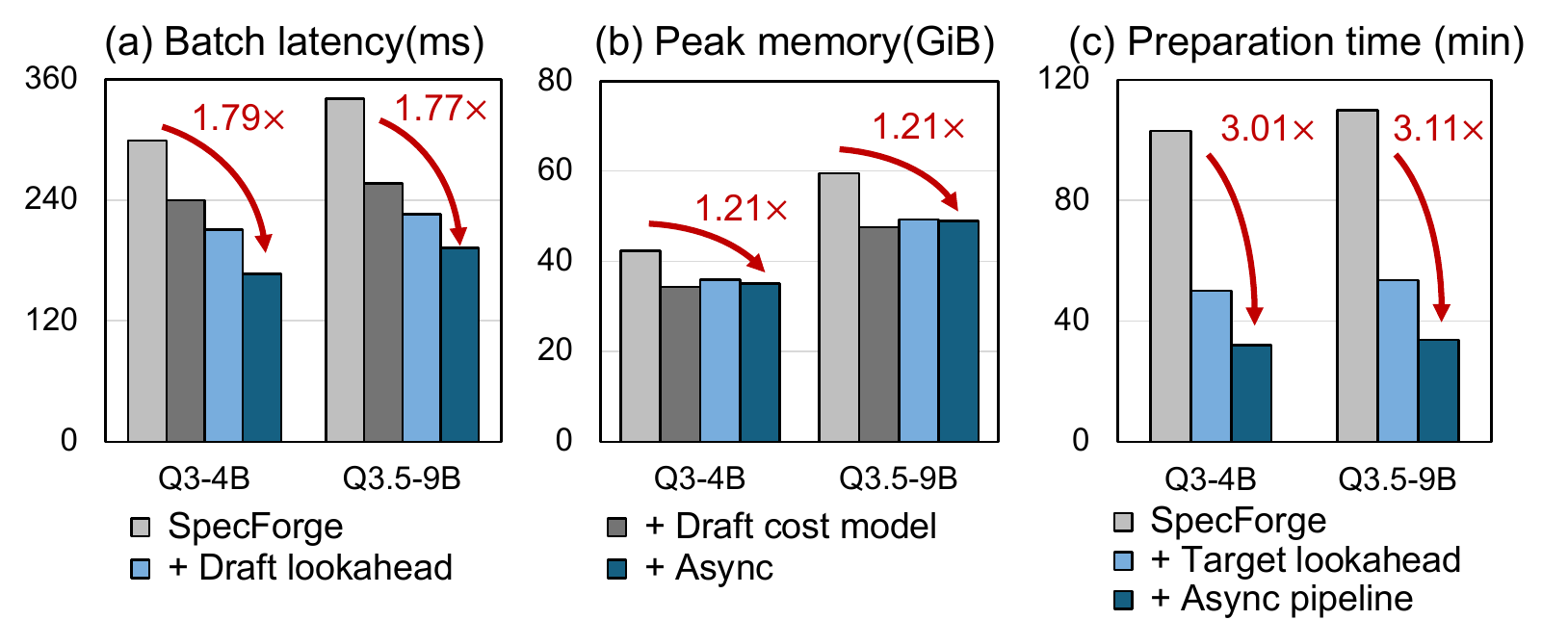}
    \vspace{-20pt}
    \caption{Framework efficiency. (a)--(b) Batch latency and peak memory under cumulative optimizations. (c) Offline supervision preparation time.}
    \vspace{-10pt}
    \label{fig:framework-efficiency}
\end{figure}

\subsection{Framework Efficiency}
\label{sec:framework-efficiency}

\paragraph{Draft Training}
\textbf{LEGOSpec} achieves a $1.77$--$1.79\times$ speedup over SpecForge with 17.1--17.6\% lower peak memory (Fig.~\ref{fig:framework-efficiency}(a)--(b)). The draft cost model gives a $1.25$--$1.33\times$ speedup by balancing draft-worker computation; draft lookahead adds $1.14\times$, and the asynchronous  pipeline adds $1.17$--$1.26\times$ by overlapping scheduling and transfer with training.

\paragraph{Offline Supervision Preparation} Target-side lookahead scheduling and asynchronous cache writing accelerate preparation of 400K samples by $3.01\times$ on Qwen3-4B and $3.11\times$ on Qwen3.5-9B (Fig.~\ref{fig:framework-efficiency}(c)). Grouping reduces padding and balances prefill work, while asynchronous writing overlaps storage I/O with target computation.

\section{Conclusion}
In this paper, we propose \textbf{LEGOSpec}, a composable and efficient framework for speculative draft model design and training. Using \textbf{LEGOSpec}, we conduct controlled component comparisons and composition studies to identify component effects and interactions. These findings guide workload-aware backbone choices and the general-purpose \textit{DFlux} recipe. \textit{DFlux} improves acceptance over the strongest baseline by up to 34.7\% on SWE-bench and 19.7\% on Alpaca. Its SGLang integration achieves the highest throughput on Qwen3.6-35B-A3B and Gemma4-12B-it at every tested batch size. Metadata-aware scheduling and asynchronous execution accelerate draft training by up to $1.79\times$ and offline supervision preparation by up to $3.11\times$ over SpecForge.

\section*{ACKNOWLEDGMENT}
This work was sponsored by the National Natural Science Foundation of China (No. 62622406, 62674144, 62561160156) and Shanghai Rising-Star Program (No. 24QB2706200).

\bibliographystyle{IEEEtran}
\bibliography{references}

\clearpage
\appendices

\section{Controlled Component Study Details}
\label{app:component-details}

\subsection{Backbone Configurations}
Table~\ref{tab:study-backbone-structure} lists the backbone configurations used in the controlled component study. The target model is Qwen3-4B. All variants use five layers, hidden size 2560, FFN intermediate size 9728, DFlash context, and an RNN proposal of rank 256. Parameter counts cover one backbone layer, including attention, FFN, and layer normalization
\begin{table}[!ht]
\vspace{-10pt}
\centering
\caption{Backbone configurations under comparable parameter budgets.}
\label{tab:study-backbone-structure}
\footnotesize
\setlength{\tabcolsep}{3pt}
\renewcommand{\arraystretch}{1.15}
\begin{tabular*}{\columnwidth}{@{\extracolsep{\fill}}llr@{}}
\toprule
Backbone & Structural configuration & Params./layer \\
\midrule
MHA & $H_q=H_{kv}=32$, $d_h=128$ & 116.66M \\
GQA & $H_q=32$, $H_{kv}=8$, $d_h=128$ & 100.93M \\
SWA & GQA with historical window $W=512$ & 100.93M \\
GDN & \shortstack[l]{$H_k=16$, $H_v=32$;\\$d_k=d_v=128$} & 116.87M \\
MLA & \shortstack[l]{$H_q=H_{kv}=32$, $r_{kv}=512$;\\$d_{\mathrm{nope}}=128$, $d_{\mathrm{rope}}=64$, $d_v=128$} & 106.60M \\
\bottomrule
\end{tabular*}
\par\smallskip
\raggedright
$H_q,H_{kv},H_k,H_v$ denote head counts; $d_h,d_k,d_v$ denote head dimensions. For MLA, $r_{kv}$ is the latent dimension, and $d_{\mathrm{nope}},d_{\mathrm{rope}}$ are the non-RoPE and RoPE query/key dimensions.
\vspace{-10pt}
\end{table}

\subsection{Token-Level Objectives}
\label{app:objective-definitions}
Let $p_i(v)$ and $q_i(v)$ denote target and draft probabilities over vocabulary $\mathcal V$ at a valid draft position $i$, and let $y_i$ be the recorded token label. The token-level losses are
\begin{align}
\ell_i^{\mathrm{CE}} &= -\log q_i(y_i),\\
\ell_i^{\mathrm{KL}} &= \sum_{v\in\mathcal V}p_i(v)\log\frac{p_i(v)}{q_i(v)},\\
\ell_i^{\mathrm{TV}} &= \frac{1}{2}\sum_{v\in\mathcal V}|p_i(v)-q_i(v)|,\\
\ell_i^{\mathrm{NLA}} &= -\log\sum_{v\in\mathcal V}\min\{p_i(v),q_i(v)\},\\
\ell_i^{\mathrm{Mix}} &= 0.1\ell_i^{\mathrm{CE}}+0.9\ell_i^{\mathrm{TV}}.
\end{align}
NLA loss applies a negative logarithm to the distribution overlap, following the direct-acceptance loss~\cite{samarin2026lk}. The overlap equals one minus the TV loss. CE loss uses a single label, whereas KL, TV, and NLA losses require target probabilities. Mix loss combines the label signal with distribution matching. Fixed position-decay weights aggregate local losses in the reference recipe; invalid supervision positions are excluded.

\subsection{Acceptance-Aware Block Aggregation}
For a complete block with $m$ supervised draft positions, write the local target--draft overlap as $a_i$. A later position contributes to accepted length only if its preceding prefix survives verification. D-PACE~\cite{wu2026d} and D-PAL~\cite{wu2026beyond} losses use dynamic position credits to account for this dependency rather than applying only fixed position weights.

The prefix-and-suffix credit construction can be written as
\begin{align}
u_j &= \prod_{k=1}^{j}s_k, &
c_i &= \sum_{j=i}^{m}u_j,\\
\mathcal L_{\mathrm{block}} &= \sum_{i=1}^{m}\operatorname{sg}(c_i)\,\ell_i,
\end{align}
where $s_k$ is a smoothed local acceptance proxy and $\operatorname{sg}$ stops gradients through the credit. D-PACE loss uses the draft probability of the recorded label as the proxy and CE loss locally. D-PAL loss uses full-vocabulary overlap as the proxy; the variant discussed in the main text uses KL loss locally. Smoothing interpolates the local proxy with one before the prefix product. The resulting credit includes both the likelihood of reaching a position and its contribution to later accepted positions. This separates the choice of local supervision from block-level loss aggregation.

\subsection{Training Curves}\label{app:training-curves}
\begin{figure*}[ht]\centering\includegraphics[width=0.94\textwidth]{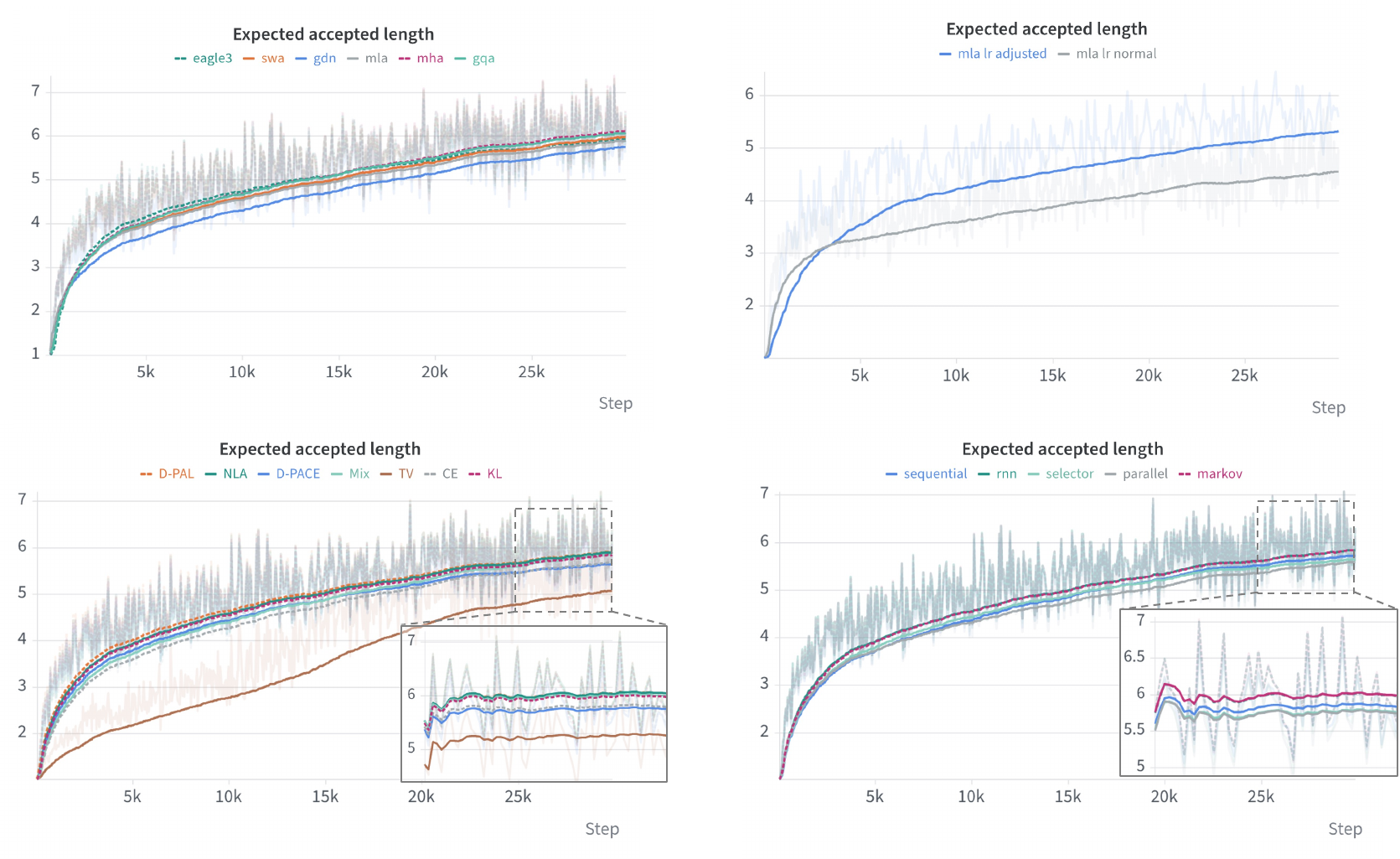}
\caption{Training curves of expected accepted length of different variants in the component study.}
\label{fig:component-training-curves}
\end{figure*}

Figure~\ref{fig:component-training-curves} shows the expected accepted length during training for the Backbone, Proposal, and Objective comparisons. During MLA backbone training, we found that using the default Muon and Adam learning rates (0.01 and 6e-4, respectively) resulted in degraded performance, possibly because MLA does not employ per-head normalization. Reducing both learning rates by an order of magnitude brought MLA performance back to a level comparable to that of other backbones. In contrast, other backbone variants and components did not exhibit similar learning-rate sensitivity, so we retained their default learning rates.

\section{Speculative Sampling Correctness}
\label{app:sampling-correctness}
Fix one prefix and abbreviate the target and proposal conditionals by $p$ and $q$. A proposal $Y\sim q$ is accepted with probability $\min\{1,p(Y)/q(Y)\}$. Define
\begin{align}
\alpha&=\sum_{v\in\mathcal V}\min\{p(v),q(v)\},\\
r(v)&=\frac{[p(v)-q(v)]_+}{1-\alpha},
\end{align}
where $[x]_+=\max\{x,0\}$ and $r$ is used only when rejection has positive probability. The probability of emitting any token $v$ is
\begin{align}
\Pr(X=v\mid \mathcal H_i)
&=\min\{p(v),q(v)\}+(1-\alpha)r(v)\\
&=p(v).
\end{align}
When $\alpha=1$, no rejection replacement is needed. Applying this identity successively at each accepted prefix, and sampling the additional target token after an entirely accepted block, recovers the target sequence distribution. The argument permits predecessor-dependent corrections and the zero-correction fallback because both are included in $q$. It assumes exact probability evaluation and the standard residual replacement rule. At $T=0$, greedy verification instead emits the target argmax at the first mismatch, using a consistent tie policy.

\section{Baseline Checkpoint Sources}
\label{sec:checkpoint-sources}
Table~\ref{tab:checkpoint-sources} lists the repositories of the baseline checkpoints; the Gemma4-12B-it DFlash2 checkpoint is trained in this work.

\begin{table*}[!t]
{\centering
\caption{Acceptance length $\tau$ at $T=1$ with matched 80K training data.}
\label{tab:matched-training-earlier}
\footnotesize
\setlength{\tabcolsep}{1.5pt}
\newcommand{\earlierdatasetheading}[1]{\makebox[0.075\textwidth][c]{#1}}
\renewcommand{\arraystretch}{1.0}
\begin{tabular*}{0.98\textwidth}{@{\extracolsep{\fill}}cccccccccccc@{}}
\toprule
 & & \multicolumn{6}{c}{Short-context workloads} & \multicolumn{3}{c}{LongBench-v2 domains} & Long coding \\
\cmidrule(lr){3-8}\cmidrule(lr){9-11}\cmidrule(lr){12-12}
Model & Method & \earlierdatasetheading{Alpaca} & \earlierdatasetheading{MT-Bench}
 & \earlierdatasetheading{GSM8K} & \earlierdatasetheading{MATH-500}
 & \earlierdatasetheading{MBPP} & \earlierdatasetheading{LCB}
 & \earlierdatasetheading{MDQA} & \earlierdatasetheading{LICL}
 & \earlierdatasetheading{DHU} & \earlierdatasetheading{SWE-bench} \\
\midrule
\multirow{4}{*}{Qwen3-4B} & DFlash & 2.66 & 3.25 & 5.25 & 4.90 & 4.14 & 3.45 & 2.15 & 2.12 & 2.50 & 1.98 \\
 & DFlash2 & 2.95 & 3.55 & 5.61 & 5.30 & 4.53 & 3.81 & 2.69 & 2.73 & 3.06 & 2.30 \\
 & DSpark & 2.99 & 3.63 & 5.73 & 5.45 & 4.67 & 4.01 & 2.32 & 2.35 & 2.68 & 2.15 \\
 & \textbf{DFlux} & \textbf{3.18} & \textbf{3.77} & \textbf{6.02} & \textbf{5.71} & \textbf{4.80} & \textbf{4.11} & \textbf{3.06} & \textbf{3.14} & \textbf{3.24} & \textbf{2.75} \\
\midrule
\multirow{4}{*}{Qwen3.5-9B} & DFlash & 2.71 & 3.08 & 5.44 & 5.04 & 4.48 & 3.34 & 2.42 & 2.37 & 2.87 & 2.18 \\
 & DFlash2 & 3.05 & 3.48 & 5.85 & 5.42 & 4.92 & 3.75 & 2.68 & 2.63 & 3.34 & 2.42 \\
 & DSpark & 3.06 & 3.51 & 5.98 & 5.58 & 5.04 & 3.86 & 2.60 & 2.58 & 3.19 & 2.35 \\
 & \textbf{DFlux} & \textbf{3.09} & \textbf{3.65} & \textbf{6.15} & \textbf{5.80} & \textbf{5.29} & \textbf{4.01} & \textbf{2.96} & \textbf{2.89} & \textbf{3.60} & \textbf{2.96} \\
\bottomrule
\end{tabular*}
\par}
\par\vspace{20pt}
\noindent
\begin{minipage}{\columnwidth}
\centering
\caption{Hugging Face repositories for baseline drafts. The Gemma4-12B-it DFlash2 baseline is trained in this work.}
\label{tab:checkpoint-sources}
\footnotesize
\setlength{\tabcolsep}{2pt}
\renewcommand{\arraystretch}{1.08}
\begin{tabular*}{\columnwidth}{@{\extracolsep{\fill}}ll@{}}
\toprule
Method & Hugging Face repo \\
\midrule
\multicolumn{2}{l}{\textbf{Qwen3-4B}} \\
\addlinespace[1pt]
DFlash & \texttt{deepseek-ai/dflash\_qwen3\_4b\_block7} \\
DFlash2 & \texttt{RedHatAI/Qwen3-4B-speculator.dflash2} \\
DSpark & \texttt{deepseek-ai/dspark\_qwen3\_4b\_block7} \\
DLite & \texttt{SJTU-DAI-Sys/Qwen3-4B-DLite} \\
\midrule
\multicolumn{2}{l}{\textbf{Qwen3.6-35B-A3B}} \\
\addlinespace[1pt]
DFlash & \texttt{z-lab/Qwen3.6-35B-A3B-DFlash} \\
DFlash2 & \texttt{incoai/Qwen3.6-35B-A3B-DFlash2} \\
DSpark & \texttt{RedHatAI/Qwen3.6-35B-A3B-speculator.dspark} \\
\midrule
\multicolumn{2}{l}{\textbf{Gemma4-12B-it}} \\
\addlinespace[1pt]
DFlash & \texttt{deepseek-ai/dflash\_gemma4\_12b\_block7} \\
DFlash2 & \texttt{aaaalex111/gemma4-12b-it-DFlash2} \\
DSpark & \texttt{deepseek-ai/dspark\_gemma4\_12b\_block7} \\
\bottomrule
\end{tabular*}

\end{minipage}\hfill\mbox{}
\end{table*}

\section{Comparison under Matched Training Conditions}
\label{app:matched-training}
Table~\ref{tab:matched-training-earlier} provides the detailed acceptance results under matched 80K training. 

\end{document}